\documentclass[runningheads]{llncs}
\usepackage{graphicx}
\usepackage{comment}
\usepackage{amsmath,amssymb} % define this before the line numbering.
\usepackage{color}
\usepackage{subfigure}
\usepackage[colorlinks,linkcolor=black]{hyperref}
\usepackage{booktabs}
\usepackage{bbding}

\newcommand{\repeatthanks}{\textsuperscript{\thefootnote}}

\begin{document}
% \linenumbers
\pagestyle{headings}
\mainmatter
\def\ECCVSubNumber{79}  % Insert your submission number here
\title{BGGAN: Bokeh-Glass Generative Adversarial Network for Rendering Realistic Bokeh} % Replace with your title

%\begin{comment}
\titlerunning{ECCV-20 submission ID \ECCVSubNumber} 
\authorrunning{ECCV-20 submission ID \ECCVSubNumber} 
\author{Anonymous ECCV submission}
\institute{Paper ID \ECCVSubNumber}
%\end{comment}
% \begin{comment}
\titlerunning{BGGAN for Rendering Realistic Bokeh}
\author{Ming Qian\inst{1}\thanks{The work was done when Ming Qian was an intern at AiRiA.} \and
Congyu Qiao\inst{2} \and
Jiamin Lin\inst{3} \and
Zhenyu Guo\inst{5,4}\and
Chenghua Li\inst{4,6}\Envelope \thanks{corresponding author} \and
Cong Leng \inst{4,6} \and
Jian Cheng \inst{4,6}\Envelope \repeatthanks
}
\authorrunning{Ming Qian, Congyu Qiao and Jiamin Lin et al.}
\institute{Nanjing University of Information Science \& Technology, China \and
Southeast University, China \and
Nanjing University of Science \& Technology, China\and
Institute of Automation, Chinese Academy of Sciences, Beijing 100190, China\and 
School of Artificial Intelligence, University of Chinese Academy of Sciences, Beijing 100049, China \and
Nanjing Artificial Intelligence Chip Research, Institute of Automation, Chinese Academy of Sciences (AiRiA)\\
\email{mingqian@nuist.edu.cn}, 
\email{\{cyqiao,linjiamin\}@njust.edu.cn},
\email{\{guozhenyu2019,lichenghua2014\}@ia.ac.cn},
\email{lengcong@airia.cn},
\email{jcheng@nlpr.ia.ac.cn}
}
% \end{comment}
%******************

\maketitle

\begin{abstract}
A photo captured with bokeh effect often means objects in focus are sharp while the out-of-focus areas are all blurred. DSLR can easily render this kind of effect naturally. However, due to the limitation of sensors, smartphones cannot capture images with depth-of-field effects directly. In this paper, we propose a novel generator called Glass-Net, which generates bokeh images not relying on complex hardware. Meanwhile, the GAN-based method and perceptual loss are combined for rendering a realistic bokeh effect in the stage of finetuning the model. Moreover, Instance Normalization(IN) is reimplemented in our network, which ensures our tflite model with IN can be accelerated on smartphone GPU. Experiments show that our method is able to render a high-quality bokeh effect and process one $1024 \times 1536$ pixel image in 1.9 seconds on all smartphone chipsets. This approach ranked First in AIM 2020 Rendering Realistic Bokeh Challenge Track 1 \& Track 2.

\keywords{Bokeh, Depth-of-field, Smartphone GPU, GAN}
\end{abstract}

\section{Introduction}
\par In photography, bokeh is considered one of the most important aesthetic standard when we need to blur the out-of-focus parts of an image produced by a camera lens. It occurs in the scene which lies outside the depth of field. Photographers sometimes deliberately adopt a shallow focus technique to create images with prominent out-of-focus regions\cite{davis2008practical} when using an SLR with a wide aperture lens. However, we can hardly take a picture with a bokeh effect using a smartphone with a monocular camera, because it is difficult to equip the little smartphone with too many sensors. Hence, it is an excellent way to render a synthetic bokeh effect on mobile devices at the level of software.
\par Synthetic bokeh effect rendering has developed for several years. Early works related to it only focus on portrait photos \cite{shen2016automatic,shen2016deep,zhu2017fast}. They usually segment the person from the image by using the semantic segmentation method and then blur the rest areas. The disadvantages of these methods are very obvious. Their datasets do not cover the images in a wide scene which are very necessary for most photographers.
\par In recent years smartphones have adopted various hardware to promote the realization of synthetic bokeh effect rendering. A method is described in detail on how to render a bokeh effect in Google Pixel devices\cite{wadhwa2018synthetic}. In this work, they synthesize depth map by using the dual-pixel autofocus system. Plus, iPhone 7+ employs the dual-lens to estimate the depth of the scene. But these methods rely on special or expensive hardware, and these approaches may not suitable for low-end smartphone market. In fact, it is better for the bokeh effect to be rendered directly from a shooted image.
\par EBB! dataset\cite{ignatov2020rendering} released by AIM 2020 Bokeh Effect Rendering Challenge\cite{ignatov2020aim_bokeh} makes it possible for us to explore some new methods about Bokeh. The dataset pays more attention to synthetic bokeh effect rendering in wide scenes. In this dataset, objects in the depth-of-field area are not only portraits but also other objects, such as road signs, vehicles, flowers, and so on. It brings many challenges but more possibilities to render the bokeh effect. Many methods\cite{purohit2019depth,dutta2020depth,ignatov2019aim} have been proposed experimented on EBB! Dataset. Most of these methods are based on some priori knowledge including salient region detection and depth estimation method. In \cite{dutta2020depth}, they propose a method based on the depth estimation method megaDepth\cite{li2018megadepth}, which is a well-known algorithm for its outstanding performance of depth estimation on a single image. Then they design an efficient algorithm to blur the regions out-of-focus by Gaussian blur kernel. A DDDF architecture is proposed in \cite{purohit2019depth}, which uses both salient region segmentation \cite{hou2017deeply} and depth estimation \cite{li2018megadepth} as its priori knowledge. Aside from these, PyNET method\cite{ignatov2020rendering} also employs the depth map produced by MegaDepth. Furthermore, they point out that depth maps can help improve the visual results despite the prior cannot increase $PSNR$ and $SSIM$, which are commonly used as quantitative indicators. From the methods mentioned above, we conclude that adding some priori knowledge like a salient region detection map or depth estimation map can improve the visual effect of the final generated image to some degree. But from another angle, those methods are relied on the priors heavily, meaning that when the priors do not work in some scenes, the final synthetic image will result in unknown problems. Plus, the pre-process of these prior methods is also time-consuming. Whenever we want to inference a new picture on our device, these models should run at first.
\par Generative Adversarial Networks(GANs)\cite{goodfellow2014generative} are well known for the ability to preserve texture details in an image, generate a more realistic image, and fool people in perception. Recent years has witnessed that a large number of image-to-image translation tasks are completed by GANs, such as image deblurring\cite{kupyn2018deblurgan}, image super-resolution\cite{ma2020structure}, style transferring\cite{karras2020analyzing}, product photo generation\cite{bousmalis2017unsupervised} and so on. These works inspire us to consider bokeh as a subtask of image-to-image translation.
\par In this paper, we describe a novel approach to the bokeh problem, which is different from the existing algorithms that rely on various priori knowledge. We don’t need the assistance of any other pre-trained models or datasets. What we need for training are only pairs of narrow-aperture images and shallow depth-of-field images. This strategy shortens the inference time compared with the methods which are based on priori knowledge. In addition, we employ a GAN-based method for finetuning our model, which turns out to be effective to improve the quality of visual effect in our work. To the best of our knowledge, we are the first to introduce the GAN method to image synthetic bokeh effect rendering. Finally, the proposed approach is  independent of hardware devices and can perform efficient synthetic bokeh effect rendering on various devices.
\par Our main contributions are:

\begin{itemize}
\item[1)] The first GAN-based method solves the synthetic bokeh effect rendering problem, and the visual effect has been greatly improved: the spot effect similar to that of large aperture SLR and the gradual blurring effect has been obtained. 
\item[2)] Compared with the baseline, a novel generator architecture not only keeps the accuracy but also significantly improves the speed.
\item[3)] IN is re-realized by GPU supported ops on tflite, that ensures our tflite model with IN can be accelerated on smartphone GPU, and process one $1024 \times 1536$ pixel image less than 1.9 seconds on all smartphone chipsets.
\end{itemize}

\section{Proposed Method}
We take Glass-Net and Multi-receptive-field Discriminator to construct our BGGAN. Glass-Net is an end-to-end network that takes an image as an input and produces the result with bokeh effect. Multi-receptive-field Discriminator refines the images generated by Glass-Net to make the final outputs cater to human perception. In addition, we use operators supported by TensorFlow Lite framework to reimplement IN to make sure all our model operators to compute on smartphone GPU.
\subsection{Glass-Net}
\par
\begin{figure}[!htbp]
    \centering
    \includegraphics[scale=0.4]{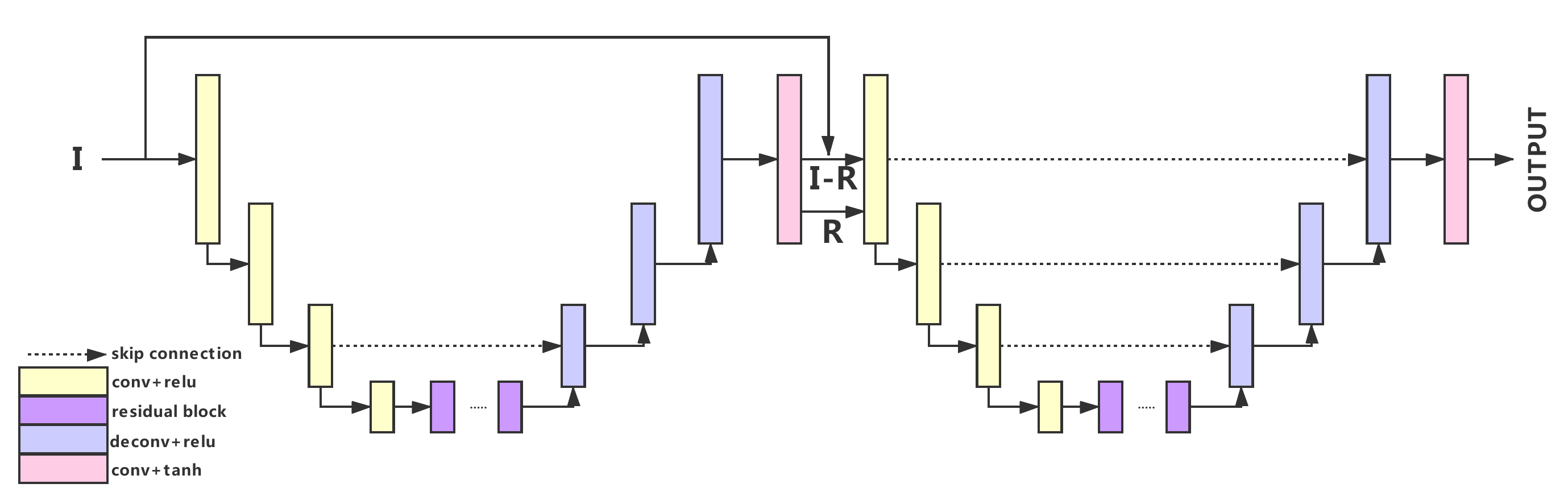}
    \caption{Demonstration for Glass-Net}
    \label{fig:glassnet}
\end{figure}

 The generator is given a name Glass-Net for the network's shape is similar to a pair of glasses. The Glass-Net is a two-stage network. Glass-Net is illustrated in Fig.\ref{fig:glassnet}. In the first stage, the network learns the mapping from the image without bokeh to the residual of the input image and ground truth. The relationship between input image $I$, ground truth $O$, and residual $R$ can be represented as $R=I-O$, so we assume $I-R$ as the rough bokeh result. Therefore, the second stage of Glass-Net plays the role of refining rough bokeh results to generate realistic bokeh effects.

\par In Fig.\ref{fig:glassnet}, we can find that the first stage of the network and the second stage of the network have the almost same structure. Both of them adopt the encoder-decoder structure. The encoder blocks consist of three down-sampling layers implemented by convolution with stride 2. We use 9 residual blocks~\cite{He2016Deep} to transform features obtained from encoder blocks and each residual block sequentially connects conv / ReLU / instancenorm / conv / ReLU layers. In addition, there is an additional connection between the input and the output of the residual block. The feature maps transformed by residual blocks are sent to decoder blocks. Decoder blocks are implemented by three transposed convolutional layers with stride 2. Convolutional layers of encoder blocks and transposed convolutional layers of decoder blocks are all activated by ReLU. The output layers of the two-stage Glass-Net are realized by convolution with stride 1 followed by tanh. Skip-connections are applied between the convolutional layer and its mirrored transposed convolutional layer to compensate the details of the output image and the operation of skip connection is concatenation on the channel. This kind of connection is also used by U-Net~\cite{ronneberger2015u} structure. Experiments proved that skip connections can convey high-frequency information so that the in-focus areas of output images will not be blurred. The number of basic channels in the first stage is 16 while the number is 32 in the second stage. The maximum number of channels in the first stage is 128 and the number of stage two is 256.

\subsection{Multi-receptive-field Discriminator}
\par
\begin{figure}[!htb]
    \centering
    \includegraphics[scale=0.4]{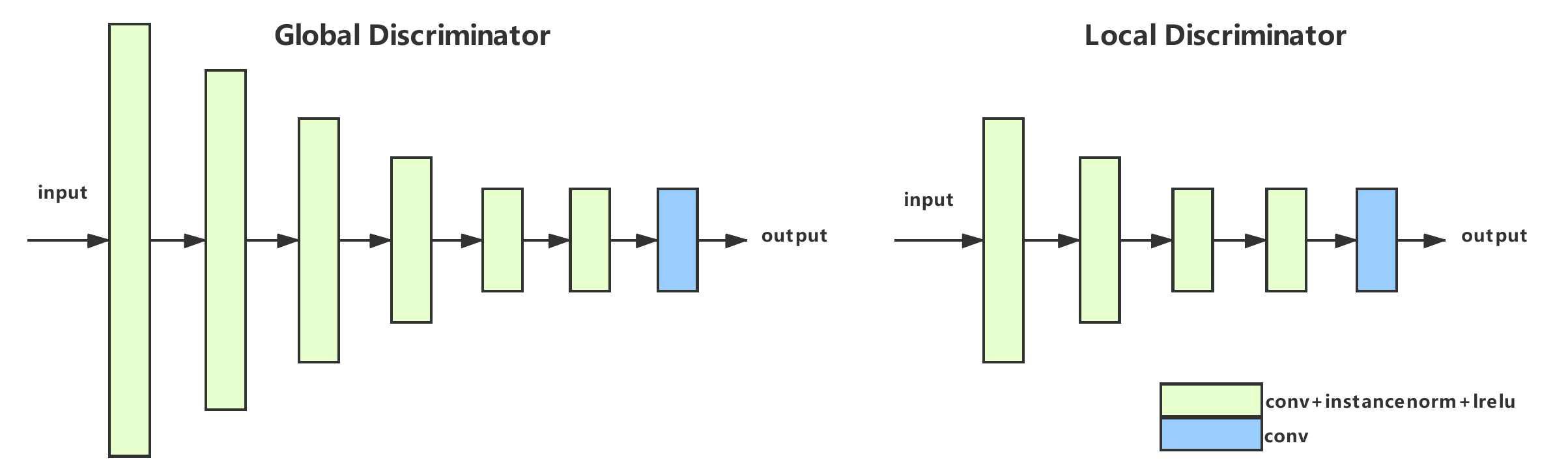}
    \caption{Demostration for Multi-receptive-field Discriminator}
    \label{fig:discriminator}
\end{figure}

 To generate more realistic bokeh images, WGAN-GP\cite{gulrajani2017improved} is used as a significant strategy in our network. Fig.\ref{fig:discriminator} illustrates the schematic diagram of Multi-receptive-field Discriminator. An additional gradient penalty in the loss function enables the finetuning process more stable and easier to produce results with higher perceptual quality.
\par Besides, the PatchGAN \cite{isola2017image}idea proposed by Isola \textit{etal.} inspires us in building the discriminator, which supervises the differences between the generated image and the ground truth on patches of size $ 70 \times 70 $. In fact, the size of the patches which the discriminator operates on coincides with that of the discriminator's receptive field. And by modifying the depth of PatchGAN discriminator, we can enable the network to pay attention to different sizes of patch details around the same pixel. Correspondingly, our generator will also improve its own performance on some details to counteract the discriminator. Therefore, we decide to apply combined PatchGAN discriminators with different receptive fields as a multi-receptive-field discriminator in the structure of the adversarial part.
\par The loss function is also important for finetuning the Glass-Net. Aside from 
$ L_{1} $  reconstruction loss and negative SSIM loss $ L_{SSIM} $, the perceptual loss which computes the Euclidean loss on the feature maps the relu5\_4 layer layer of the VGG19 $ L_{VGG} $ and the adversarial loss $ L_{adv} $ for the Glass-Net generator will also be optimized in this stage. The perceptual loss is as follow:

\begin{equation}
    \begin{aligned}
      L_{VGG}=\frac{1}{HWC}\sum_{i=1}^{H}\sum_{j=1}^{W}\sum_{k=1}^{C}\left \| F(G(I)_{i,j,k}) - F(C_{i,j,k}) \right \|_{1},
    \end{aligned}
\end{equation} 

where $ F(\cdot) $ denotes feature maps of the 34-th layer of the VGG
network which is pre-trained on ImageNet,$ G(I_{i,j}) $
denotes the image Glass-Net produces and C denotes the ground truth.
\par And the adversarial loss for the Glass-Net generator can be expressed as:

\begin{equation}
    \begin{aligned}
      L_{adv}=-\frac{1}{HW}\sum_{i=1}^{H}\sum_{j=1}^{W} D(G(I)_{i,j})
    \end{aligned}
\end{equation}

\par where $ D(\cdot) $ denotes the output of the discriminator.
\par We incorporate the four loss functions into a hybrid loss $ L_{hybrid} $ for finetuning our generator Glass-Net with appropriate weights. The hybrid loss is defined as

\begin{equation}
    \begin{aligned}
      L_{hybrid} = 0.5 \times L_{1} + 0.05 \times L_{SSIM} + 0.1 \times L_{VGG} + L_{adv}
    \end{aligned}
\end{equation}

$ L_{adv} $ is endowed with a larger factor because we want GAN to play a leading role in the optimizing progress. Compared to the perceptual loss in PyNET, we enhance the effect of $ L_{VGG} $ and weaken $ L_{SSIM} $, which contributes to reducing the disorder of out-of-focus areas.

\subsection{Reimplemented Instance Normalization}
\par Instance Normalization(IN)\cite{ulyanov2016instance} is an effective trick for image-to-image translation tasks. If we remove IN from Glass-Net, the generated images will not be as attractive as the images produced by BGGAN with IN. In \cite{ignatov2020rendering}, they discussed the original implemented IN are still not supported adequately by the TensorFlow Lite framework, thus using IN will increase the inference time and memory consumption due to additional CPU-GPU synchronization. To solve this problem, we use operators supported by tflite framework to reimplement IN to make sure all our model ops compute on smartphone GPU.
\par To begin with, we should find out which operator cannot be accelerated by tflite framework on smartphone GPUs. As we know, IN is defined as follow:

\begin{equation}
    \begin{aligned}
      y_{tijk}=\frac{x_{tijk}-\mu_{ti}}{\sqrt{\sigma_{ti}^{2}+\epsilon}}
    \end{aligned}
\end{equation}

\begin{equation}
    \begin{aligned}
      \mu_{ti}=\frac{1}{HW}\sum_{i=1}^{H}\sum_{j=1}^{W}x_{tijk}
    \end{aligned}
\end{equation}

\begin{equation}
    \begin{aligned}
      \sigma_{ti}^{2}=\frac{1}{HW}\sum_{i=1}^{H}\sum_{j=1}^{W}(x_{tijk}-\mu_{ti})^{2}
    \end{aligned}
\end{equation}

\par The challenge of the IN's implementation is computing $ \mu_{ti} $ and $ \sigma_{ti} $. Usually, both of them can be worked out through the operation `tf.nn.moment' in the program code. Alternatively, We can use the command `tf.reduce\_mean' to compute $ \mu_{ti} $ first and calculate $\sigma_{ti}$ later with $\mu_{ti}$. However, according to the guidelines of Tflite,  `tf.nn.moment' and `tf.reduce\_mean'  with the parameter `axis' are not supported on smartphone GPU. Different from the common method, we ingeniously adopt `tf.nn.avg\_pool2d' to calculate $\mu_{ti}$ due to the constant size of the feature maps at each layer.

\section{Experiments}

\subsection{Dataset}
EBB! dataset\cite{ignatov2020rendering} was released by AIM 2020 Bokeh Effect Rendering Challenge. The dataset consists of 5K shallow/wide depth-of-field image pairs. 4600 image pairs are training dataset, the number of validation data and test data are 200 respectively. In each photo pair, the image without bokeh effect is captured with a narrow aperture (f/16), while the corresponding bokeh image is shot using the highest aperture (f/1.8). The photos are captured in a variety of places with automatic mode. Though the dataset has kinds of scenes, we find that EBB! dataset has some poor-aligned image pairs. So we manually clean the train data of EBB! dataset to 4464 images as to make sure our model learn a better mapping from the original image to bokeh image.

\begin{figure}[!ht]
	\centering
	\centering
	\includegraphics[width=1\linewidth]{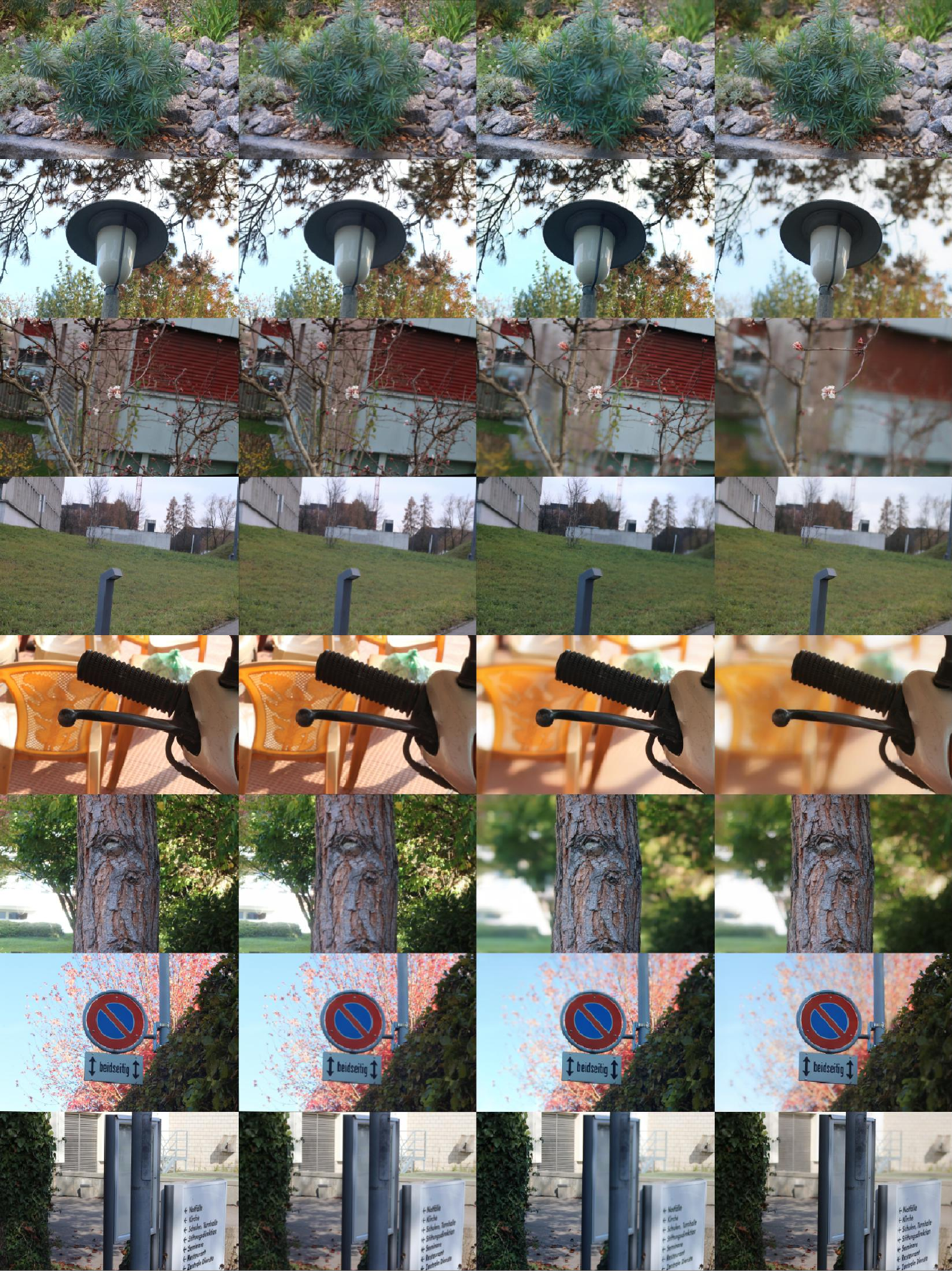}
	\caption{From left to right: input images, results of Dutta et al., results of PyNET, our results. Some pictures show that the method of Dutta et al. and PyNET fail to separate front scenes from back scenes due to bad depth maps obtained from Megadepth.}
    \label{fig:Qualitative}
\end{figure}

\subsection{Training Details}
TensorFlow is used to implement our algorithm. All experiments run on the server with NVIDIA TITAN RTX and Intel Xeon Gold 5220 CPU. In training, we randomly cropped the images to $1408\times1024$ sized patches as inputs. The batch size was set to 1 limited by memory, and the number of epochs was set to 60 in two training stages. In order to accelerate the convergence of the model, we apply a two-stage training strategy. For the first training stage, we use L1 loss combined with SSIM loss to train the network roughly. And in the second stage, we joint L1 loss, Adversarial loss, perceptual loss, and SSIM loss to finetune the network. The learning rate of the discriminator and generator in the two training stages are both set to $1e-4$. In addition, we adopt Adam as optimizer for training. It's noted that $\beta _{1}$ and $\beta _{2}$ are set to 0 and 0.9 for Adam.
\subsection{Quantitative Evaluation}
\par Our model is proposed to participate in AIM 2020 Bokeh Effect challenge which encourages participants to provide a solution to generate bokeh effect in bokeh-free images. At the beginning of the competition, the results of us are compared with the methods proposed by other teams to test quantitatively on the common metrics $ PSNR $ , $ SSIM $ only for reference. Then, the organizer conducts a user study in which all users evaluate each method by selecting one of the five quality levels(0 - mostly different from the ground truth, 5 - almost identical). The scores are then averaged per each approach to obtain the final Mean Option Scores $MOS$, which is more convincing 
than $PSNR$ and $SSIM$. Also, in order to 
further increase the persuasiveness, PyNet \cite{ignatov2020rendering} and two excellent methods \cite{dutta2020depth} \cite{ignatov2019aim} in AIM 2019 Bokeh Effect challenge are used for contrast. Table.\ref{Table:Quantitative} shows the results. Evaluated on the test dataset, the $ PSNR $ of our proposed method ranks second and the $MOS$ ranks first.

% \begin{center}
\begin{table}[]
\centering
\begin{tabular}{lccc}
\hline
team/method                                        & PSNR    & SSIM    & MOS \\ \hline
Dutta et al.\cite{dutta2020depth}                  & 22.14   & 0.8778 & - \\
xuehuapiaopiao-team                                & 22.97   & 0.8758  & 4.1\\
Terminator                                         & 23.04   & 0.8756  & 4.1\\
CET\_CVLab                                          & 23.05   & 0.8591 &  3.2\\
Team Horizon                                       & 23.27   & 0.8818  & 3.2 \\
PyNET\cite{ignatov2020rendering}                   & 23.28   & 0.8780 &  4.1\\
Zheng et al.\cite{ignatov2019aim}                  & 23.44   & 0.8874 &  - \\
AIA-Smart                                          & 23.55   & 0.8829  & 3.8 \\
IPCV\_IITM                                         & 23.77   & 0.8866 &  2.5 \\
Ours                                               & 23.58   & 0.8770  &  \textbf{4.2} \\ \hline
\end{tabular}
\caption{The results on the EBB! test subset obtained with different solutions.}
\label{Table:Quantitative}
\end{table}
% \end{center}
\subsection{Qualitative Evaluation}
\par We analyzed qualitatively by comparing our results with PyNET\cite{ignatov2020rendering} and the method proposed by Dutta et al.\cite{dutta2020depth}. And some visual results are shown in Fig.\ref{fig:Qualitative}. It should be highlighted that we did not certain the Megadepth pretrained model used by PyNET, so we can only choose the official pretrained weights downloaded from \href{http://www.cs.cornell.edu/projects/megadepth/dataset/models/best_generalization_net_G.pth}{MegaDepth}. It was introduced in the official website that the pretrained weights have well-down generalization ability to completely unknown scenes. And the method of Dutta et al. is reproduced by us with the same Megadepth weights mentioned above.

\par It's easy to find that BGGAN generates the most natural bokeh effect among the three approaches. The objects of interest in the bokeh images produced by our network are clear while the other two results are a bit fuzzy. What's more, our network can separate front scenes from back scenes well while the other two solutions can't do so. This is because PyNET and the approach of Dutta et al. rely on the depth map generated by Megadepth very much. Once depth maps cannot provide accurate depth information, the results produced by the two methods will become unattractive. Our full test results will be released on \href{https://github.com/qianmingduowan/AIM2020-bokeh-BGNet}{github.com/qianmingduowan/AIM2020-bokeh-BGNet}.

\subsection{Ablation Study and Analysis}

\subsubsection{Effect of IN}
\par
\begin{figure}[!ht]
	\centering
	\centering
	\includegraphics[width=1\linewidth]{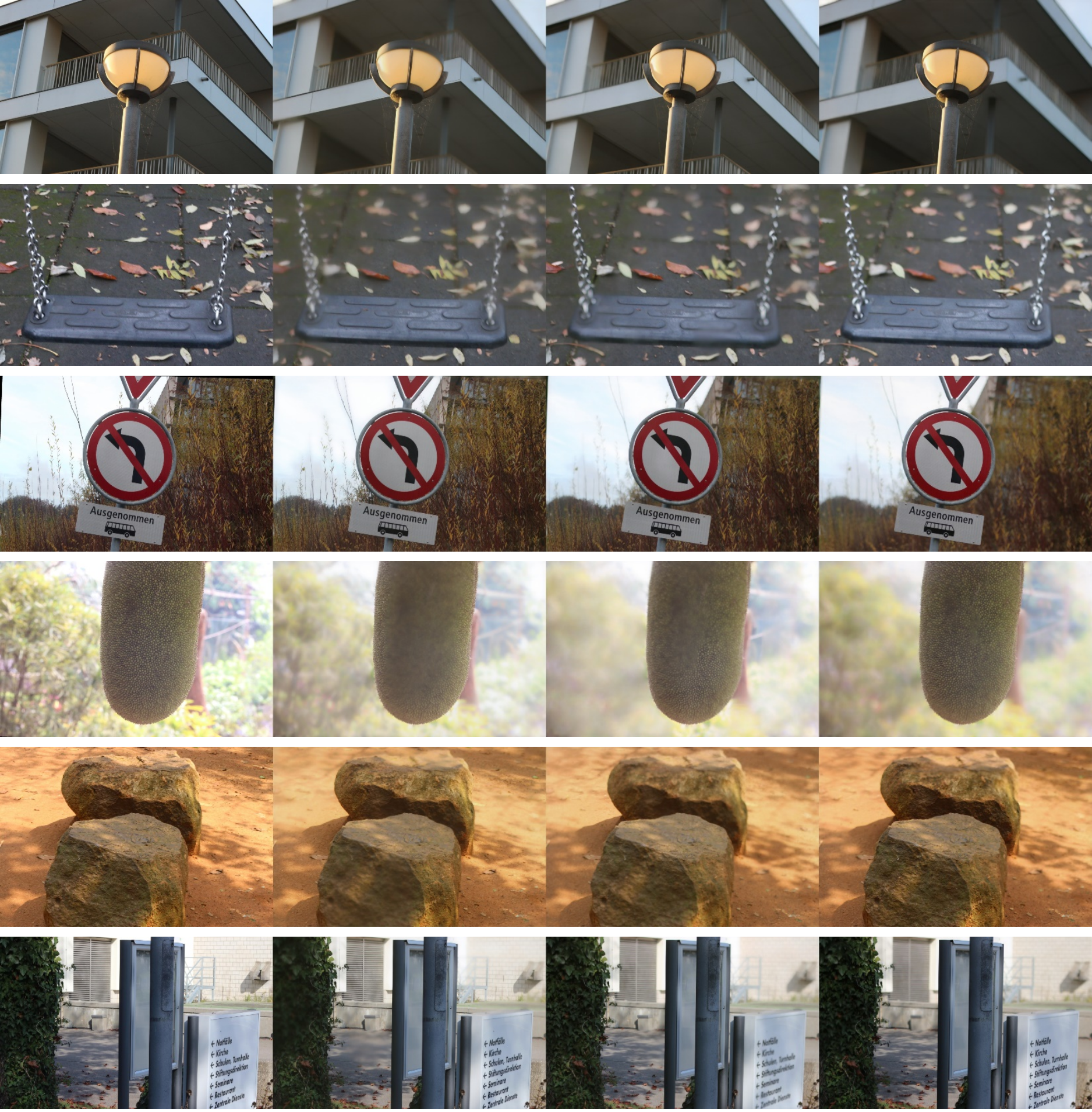}
	\caption{Effects of different normalization. From left to right: original image, results of BGGAN with batch normalization,results of BGGAN without normalization, results of BGGAN with IN.}
    \label{fig:compare_norm}
\end{figure}

 To verify the effectiveness of IN concerning the improvement in visual quality, we compare three variants of the generator: Glass-Net with IN, Glass-Net without Normalization, Glass-Net with batch normalization(BN) are investigated. Besides, we maintain the same training strategy and the same structure of the discriminator to ensure the effectiveness of the ablation study. We show our results in Fig.\ref{fig:compare_norm}. From the illustration, we can find that generator with BN has the poorest Visual result. A possible explanation is that BN considers more about the characteristics of the whole training dataset rather than individual characteristics. In contrast, Glass-Net without Normalization performs better than the model with BN. Glass-Net with IN has the best performance in qualitative research.

\subsubsection{Effect of Discriminator} 
\par To prove the effectiveness of our discriminator, we conduct the comparative experiment between the results of Glass-Net with the discriminator and the model without it at the stage of finetuning. Correspondingly, the loss function of Glass-Net without the discriminator does not possess $ L_{adv} $ and is defined as:

\begin{equation}
    \begin{aligned}
      L_{hybrid} = 0.5 \times L_{1} + 0.05 \times L_{SSIM} + 0.1 \times L_{VGG}
    \end{aligned}
\end{equation}

\begin{figure}[!htb]
	\centering
	\centering
	\includegraphics[width=1\linewidth]{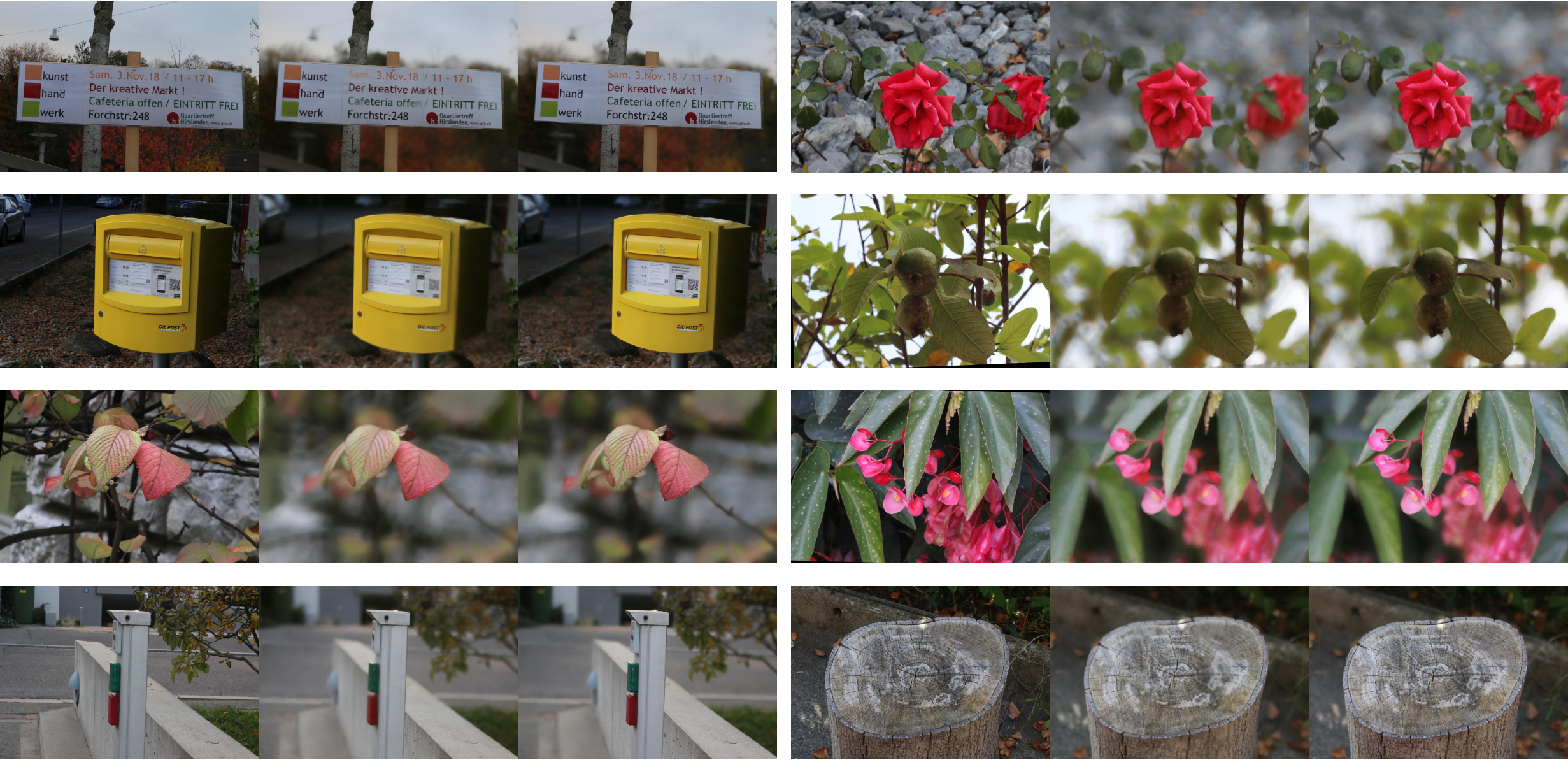}
	\caption{Comparison between the results of Glass-Net with the discriminator and the model without it. For the single photo from left to right: the input image, the result of Glass-Net without the discriminator and the result of Glass-Net with the discriminator.}
    \label{fig:effect_of_discriminator}
\end{figure}

Then the two model are trained respectively and some visual results are shown in Fig.\ref{fig:effect_of_discriminator}. By comparison, we can find out two obvious improvements in perceptual quality. The first thing that should be highlighted is that some out-of-focus areas of an image become tidier and the artifacts disappear completely. The well-distributed blur effect makes the whole image look more pleasing. Also, it should be noted that the model with the discriminator has a greater ability to segment the DoF area in photos. On the one hand, some areas in focus which Glass-Net mistakes for the out-of-focus areas are segmented correctly by the model with the discriminator. On the other hand, the boundaries between the in and out of focus areas are rendered much better.  

\subsection{Evaluation on the photos taken by the smartphone}

\par
\begin{figure}[htb]
	\centering
	\centering
	\includegraphics[width=1\linewidth]{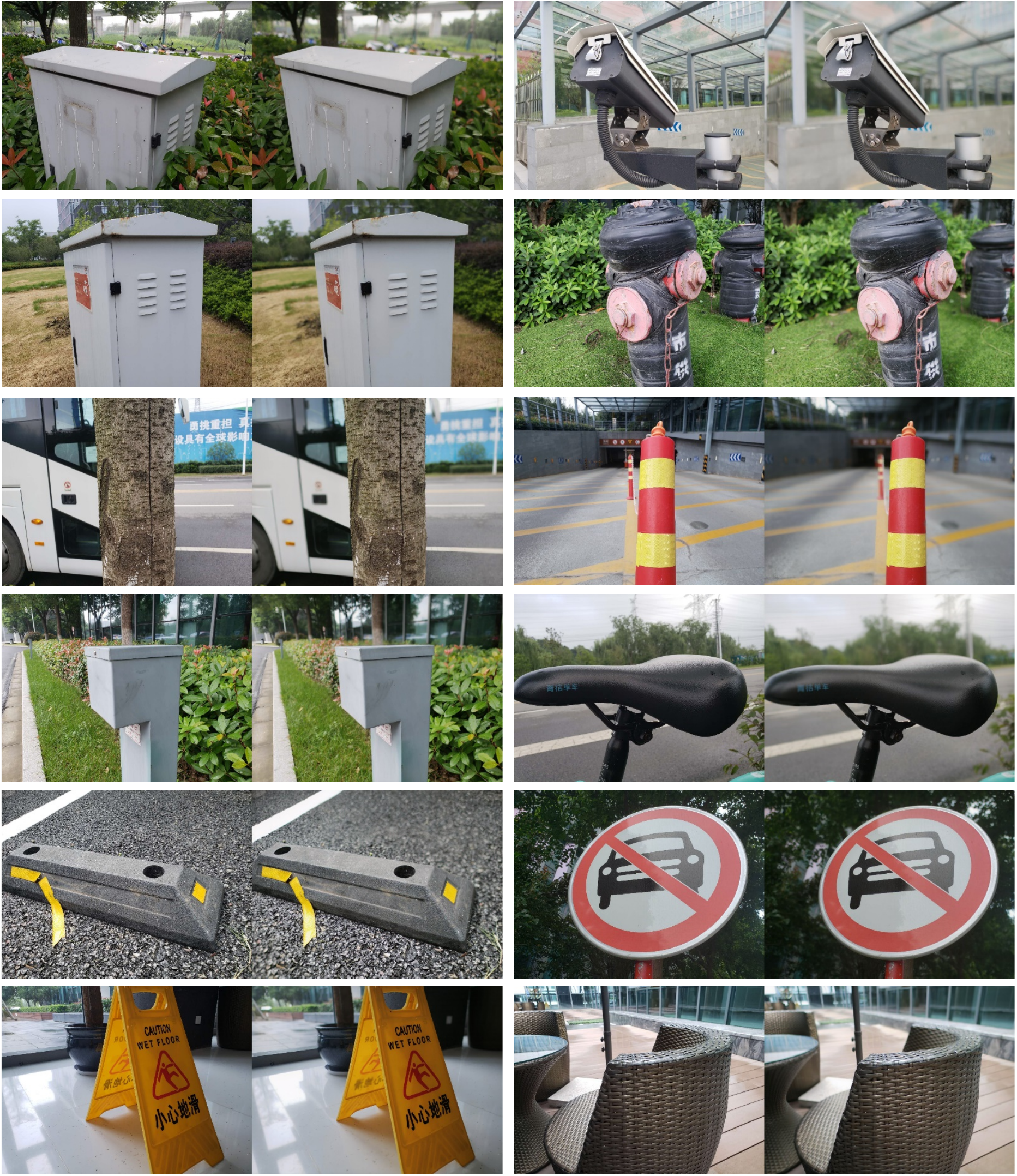}
	\caption{Images taken by Samsung S9 are on the left, and bokeh images processed by BGGAN are on the right.}
    \label{fig:sumsung}
\end{figure}

\begin{figure}[htb]
	\centering
	\centering
	\includegraphics[width=1\linewidth]{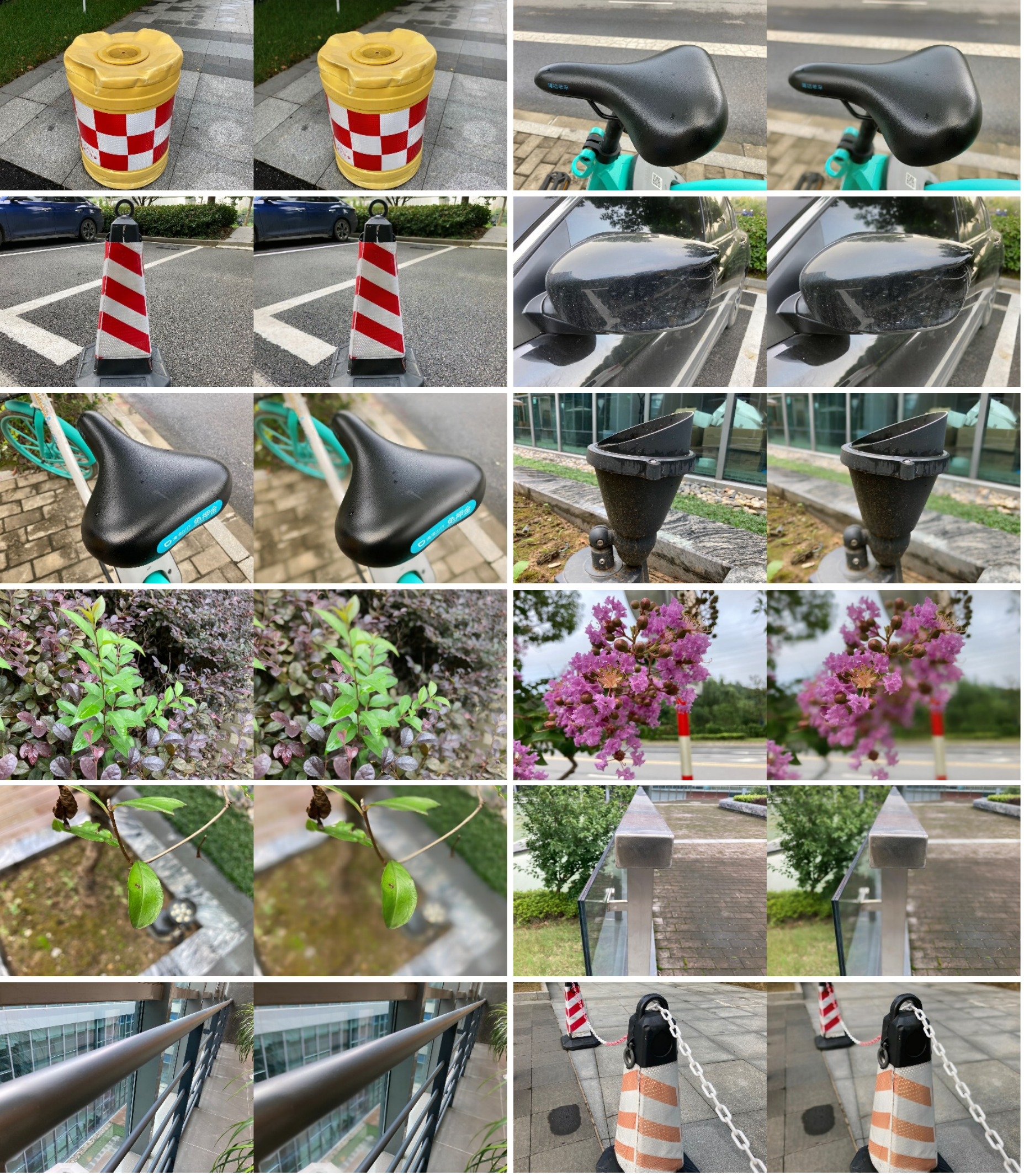}
	\caption{Images taken by iPhone SE2 are on the left, and bokeh images processed by BGGAN are on the right.}
    \label{fig:iphone}
\end{figure}

 To prove that the BGGAN is also effective in real scenarios, we take some photos of different scenes by iPhone SE2 and Samsung S9, and the photos are processed by BGGAN to get the bokeh effect. The images taken by smartphones and bokeh images processed by our network are shown in Fig.\ref{fig:iphone} and Fig.\ref{fig:sumsung} . From the samples below, we can find that even in real-world scenes, BGGAN can render realistic bokeh well, which proves our network has a strong generalization ability.

\subsection{The latency on Mobile Devices:}
\par Last,  Table.\ref{Table:GPU speed} was presented to validate the reimplemented IN on smartphone GPU which were obtained with the PRO Mode of \href{http://ai-benchmark.com/}{the AI Benchmark application}\cite{ignatov2019aim}. Our reimplemented IN can run on all kinds of SoCs of different smartphones, while the original IN only works on Exynos 9820 and Kirin 980. The reason why the original IN cannot run on Snapdragon 855 and Snapdragon 845 is that some operations in it are not supported to be accelerated by the TensorFlow Lite framework and are computed on the CPU, which results in the increase of the inference time and the great consumption memory due to additional CPU-GPU synchronization. In Table.\ref{Table:GPU speed}, we compare the average running time of processing one single image of $1024 \times 1536$ pixel, which indicates that our reimplemented IN is suitable for TensorFlow Lite framework and is of sufficient practical value. Our method was able to process one $1024 \times 1536$ pixel photo in less than 1.9 seconds on all chipsets.

\begin{table}[h!]
% \resizebox{\textwidth}{15mm}{}
\begin{tabular}{l|c|c|c|c}
Mobile Chipset                                                         & Exynos 9820 & Kirin 980 & Snapdragon 845 & Snapdragon 855 \\ \hline
GPU Model                                                              & Mali-G76    & Mali-G76  & Adreno 630     & Adreno 640     \\ \hline
CET\_CVLab                                                             & 2.0         & 3.0       & 3.0            & 3.4            \\ \hline
PyNET with original IN                                                 & 17.3        & 13.4      & -              & -              \\ \hline
BGGAN with original IN                                                 & 7.8         & 6.4       & -              & -              \\ \hline
\begin{tabular}[c]{@{}l@{}}BGGAN with \\ Reimplemented IN\end{tabular} & 1.5         & 1.4       & 1.5            & 1.8            \\ \hline
\end{tabular}
\caption{Average processing time for resolution $1024 \times 1536$ pixel obtained on several mainstream high-end mobile SoCs. In each case, the model was running directly on the corresponding GPU with OpenCL-based TensorFlow Lite GPU delegate\cite{lee2019device}.}
\label{Table:GPU speed}
\end{table}

\begin{figure}[!htbp]
	\centering
	\centering
	\includegraphics[width=1\linewidth]{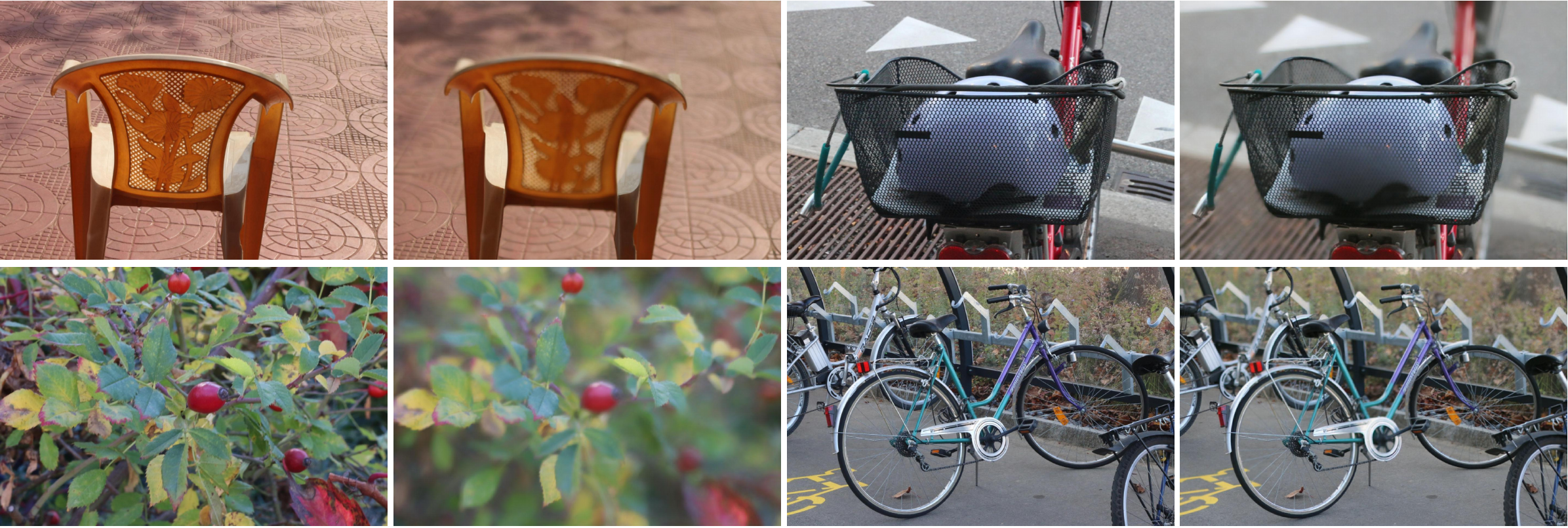}
	\caption{Some failure results. 
% 	Top Left: our network sometimes cannot render a bokeh effect for a similar color between depth-of-field and out focus areas. Top Right: fails to distinguish region with similar local textures, e.g. basket and helmet. Bottom Left: the scene is too complicated for our model to rendering realistic bokeh. Bottom Right: fine-grained information hardly to be distinguished.  
	}
    \label{fig:success}
\end{figure}

\section{Conclusions}
\par In this paper, we put forward a novel approach for the realistic bokeh effect rendering task. The proposed architecture BGGAN is the first GAN-based method to solve the problem of synthetic bokeh effect rendering. Experiments show that our approach enhances visual effects greatly compared with previous methods. In the end, we use the operators supported by tflite framework to reimplement IN to make sure that our full solution can run completely on smartphone GPU rather than partly on smartphone CPU. Thanks to the reimplemented IN, our method was able to process one $1024 \times 1536$ pixel photo in less than 1.9 seconds on all chipsets. This approach ranked First in AIM 2020 Rendering Realistic Bokeh Challenge Track 1 \& Track 2. We conclude that our method is an effective solution for the realistic bokeh effect rendering task. However, our method also has a number of limitations: BGGAN still does not work well when the color of the objects in depth-of-field is similar to that of the surrounding background. And it also happens when the scene of the whole picture is complex. The examples of visual results are represented in Fig.\ref{fig:success}. The probable reason for this phenomenon is that the model has not learned enough depth information.  But if we take some priori knowledge such as MegaDepth, it will lead to another problem: the increase of reference time. So this is a trade-off. In the future, we can make efforts in this direction. Considering that in the stage of model designing there are no good means to quantify the improvement of visual effects by the model, we can only score the picture one by one through the naked eye. How to quantify the improvement of visual quality is an essential research direction.

\section*{Acknowledgements}
This work was supported by the Advance Research Program (31511130301); National Key Research and Development Program (2017YFF0209806), and National Natural Science Foundation of China (No. 61906193; No. 61906195; No. 61702510).

\clearpage
\bibliographystyle{splncs04}
\bibliography{egbib}

\begin{thebibliography}{10}
\providecommand{\url}[1]{\texttt{#1}}
\providecommand{\urlprefix}{URL }
\providecommand{\doi}[1]{https://doi.org/#1}

\bibitem{bousmalis2017unsupervised}
Bousmalis, K., Silberman, N., Dohan, D., Erhan, D., Krishnan, D.: Unsupervised
  pixel-level domain adaptation with generative adversarial networks. In:
  Proceedings of the IEEE conference on computer vision and pattern
  recognition. pp. 3722--3731 (2017)

\bibitem{davis2008practical}
Davis, H.: Practical artistry: light \& exposure for digital photographers. "
  O'Reilly Media, Inc." (2008)

\bibitem{dutta2020depth}
Dutta, S.: Depth-aware blending of smoothed images for bokeh effect generation.
  arXiv preprint arXiv:2005.14214  (2020)

\bibitem{goodfellow2014generative}
Goodfellow, I., Pouget-Abadie, J., Mirza, M., Xu, B., Warde-Farley, D., Ozair,
  S., Courville, A., Bengio, Y.: Generative adversarial nets. In: Advances in
  neural information processing systems. pp. 2672--2680 (2014)

\bibitem{gulrajani2017improved}
Gulrajani, I., Ahmed, F., Arjovsky, M., Dumoulin, V., Courville, A.C.: Improved
  training of wasserstein gans. In: Advances in neural information processing
  systems. pp. 5767--5777 (2017)

\bibitem{He2016Deep}
He, K., Zhang, X., Ren, S., Sun, J.: Deep residual learning for image
  recognition. In: Conference on Computer Vision and Pattern Recognition. pp.
  770--778. IEEE (2016)

\bibitem{hou2017deeply}
Hou, Q., Cheng, M.M., Hu, X., Borji, A., Tu, Z., Torr, P.H.: Deeply supervised
  salient object detection with short connections. In: Proceedings of the IEEE
  Conference on Computer Vision and Pattern Recognition. pp. 3203--3212 (2017)

\bibitem{ignatov2020rendering}
Ignatov, A., Patel, J., Timofte, R.: Rendering natural camera bokeh effect with
  deep learning. In: Proceedings of the IEEE/CVF Conference on Computer Vision
  and Pattern Recognition Workshops. pp. 418--419 (2020)

\bibitem{ignatov2019aim}
Ignatov, A., Patel, J., Timofte, R., Zheng, B., Ye, X., Huang, L., Tian, X.,
  Dutta, S., Purohit, K., Kandula, P., et~al.: Aim 2019 challenge on bokeh
  effect synthesis: Methods and results. In: 2019 IEEE/CVF International
  Conference on Computer Vision Workshop (ICCVW). pp. 3591--3598. IEEE (2019)

\bibitem{ignatov2020aim_bokeh}
Ignatov, A., Timofte, R., et~al.: {AIM 2020} challenge on rendering realistic
  bokeh. In: European Conference on Computer Vision Workshops (2020)

\bibitem{isola2017image}
Isola, P., Zhu, J.Y., Zhou, T., Efros, A.A.: Image-to-image translation with
  conditional adversarial networks. In: Proceedings of the IEEE conference on
  computer vision and pattern recognition. pp. 1125--1134 (2017)

\bibitem{karras2020analyzing}
Karras, T., Laine, S., Aittala, M., Hellsten, J., Lehtinen, J., Aila, T.:
  Analyzing and improving the image quality of stylegan. In: Proceedings of the
  IEEE/CVF Conference on Computer Vision and Pattern Recognition. pp.
  8110--8119 (2020)

\bibitem{kupyn2018deblurgan}
Kupyn, O., Budzan, V., Mykhailych, M., Mishkin, D., Matas, J.: Deblurgan: Blind
  motion deblurring using conditional adversarial networks. In: Proceedings of
  the IEEE conference on computer vision and pattern recognition. pp.
  8183--8192 (2018)

\bibitem{lee2019device}
Lee, J., Chirkov, N., Ignasheva, E., Pisarchyk, Y., Shieh, M., Riccardi, F.,
  Sarokin, R., Kulik, A., Grundmann, M.: On-device neural net inference with
  mobile gpus. arXiv preprint arXiv:1907.01989  (2019)

\bibitem{li2018megadepth}
Li, Z., Snavely, N.: Megadepth: Learning single-view depth prediction from
  internet photos. In: Proceedings of the IEEE Conference on Computer Vision
  and Pattern Recognition. pp. 2041--2050 (2018)

\bibitem{ma2020structure}
Ma, C., Rao, Y., Cheng, Y., Chen, C., Lu, J., Zhou, J.: Structure-preserving
  super resolution with gradient guidance. In: Proceedings of the IEEE/CVF
  Conference on Computer Vision and Pattern Recognition. pp. 7769--7778 (2020)

\bibitem{purohit2019depth}
Purohit, K., Suin, M., Kandula, P., Ambasamudram, R.: Depth-guided dense
  dynamic filtering network for bokeh effect rendering. In: 2019 IEEE/CVF
  International Conference on Computer Vision Workshop (ICCVW). pp. 3417--3426.
  IEEE (2019)

\bibitem{ronneberger2015u}
Ronneberger, O., Fischer, P., Brox, T.: U-net: Convolutional networks for
  biomedical image segmentation. In: International Conference on Medical image
  computing and computer-assisted intervention. pp. 234--241. Springer (2015)

\bibitem{shen2016automatic}
Shen, X., Hertzmann, A., Jia, J., Paris, S., Price, B., Shechtman, E., Sachs,
  I.: Automatic portrait segmentation for image stylization. In: Computer
  Graphics Forum. vol.~35, pp. 93--102. Wiley Online Library (2016)

\bibitem{shen2016deep}
Shen, X., Tao, X., Gao, H., Zhou, C., Jia, J.: Deep automatic portrait matting.
  In: European conference on computer vision. pp. 92--107. Springer (2016)

\bibitem{ulyanov2016instance}
Ulyanov, D., Vedaldi, A., Lempitsky, V.: Instance normalization: The missing
  ingredient for fast stylization. arXiv preprint arXiv:1607.08022  (2016)

\bibitem{wadhwa2018synthetic}
Wadhwa, N., Garg, R., Jacobs, D.E., Feldman, B.E., Kanazawa, N., Carroll, R.,
  Movshovitz-Attias, Y., Barron, J.T., Pritch, Y., Levoy, M.: Synthetic
  depth-of-field with a single-camera mobile phone. ACM Transactions on
  Graphics (TOG)  \textbf{37}(4),  1--13 (2018)

\bibitem{zhu2017fast}
Zhu, B., Chen, Y., Wang, J., Liu, S., Zhang, B., Tang, M.: Fast deep matting
  for portrait animation on mobile phone. In: Proceedings of the 25th ACM
  international conference on Multimedia. pp. 297--305 (2017)

\end{thebibliography}
\end{document}